%% file: main.tex
\documentclass[journal,twoside,web]{ieeecolor}
\usepackage{tmi}
\usepackage{cite}
\usepackage{amsmath,amssymb,amsfonts}
\usepackage{algorithmic}
\usepackage{graphicx}
\usepackage{textcomp}
\usepackage{multirow}
\usepackage{booktabs}
\usepackage{makecell}
\usepackage[table]{xcolor}

\def\BibTeX{{\rm B\kern-.05em{\sc i\kern-.025em b}\kern-.08em
    T\kern-.1667em\lower.7ex\hbox{E}\kern-.125emX}}
\begin{document}
\title{Multi-Aspect Knowledge-Enhanced Medical Vision-Language Pretraining with Multi-Agent Data Generation}
\author{Xieji Li, Siyuan Yan$^\dag$, Yingsheng Liu, H. Peter Soyer, Monika Janda, Victoria Mar, and Zongyuan Ge, \IEEEmembership{Senior Member, IEEE}
\thanks{$\dag$Corresponding author: siyuan.yan@monash.edu.}
\thanks{This work was supported by the National Health and Medical Research Council (NHMRC) Centre of Research Excellence in Skin Imaging and Precision Diagnosis (Grant No. 2006551), NHMRC SYNERGY Grant: Roadmap Options for Melanoma Screening in Australia (Grant No. 2009923), and NHMRC Investigator Grant (Grant No. 2034422).}
\thanks{X.L., S.Y., Y.L., and Z.G. are with the Department of Data Science and AI, Faculty of Information Technology, Monash University, Clayton, VIC 3800, Australia.}
\thanks{V.M. is with Victorian Melanoma Service, Alfred Health, Melbourne, VIC 3004, Australia.}
\thanks{M.J. and H.P.S. are with the Frazer Institute, The University of Queensland, Dermatology Research Centre, Brisbane, Queensland, Australia, QLD 4072, Australia.}
}

\maketitle
\begin{abstract}
\input{section/1_abstract}
\end{abstract}

\section{INTRODUCTION}
\label{sec:introduction}
\input{section/2_introduction}

\section{RELATED WORK}
\input{section/3_relatedwork}

\section{METHOD}

\input{section/4_method}

\section{EXPERIMENTS}\label{sec:experiments}
\input{section/5_experiments}

\section{CONCLUSION}

\input{section/6_discussion}

\bibliographystyle{IEEEtran}
\bibliography{IEEEabrv, main.bib}

\end{document}

%% file: section/1_abstract.tex
Vision–language pretraining (VLP) has emerged as a powerful paradigm in medical image analysis, enabling representation learning from large-scale image–text pairs without relying on expensive manual annotations. However, existing methods often struggle with the noise inherent in web-collected data and the complexity of unstructured long medical texts. To address these challenges, we propose a novel VLP framework integrating a Multi-Agent data GENeration (MAGEN) system and Ontology-based Multi-Aspect Knowledge-Enhanced (O-MAKE) pretraining. First, MAGEN enhances data quality by synthesizing knowledge-enriched descriptions via a foundation model–assisted captioning and retrieval-based verification pipeline. Second, O-MAKE addresses the difficulty of learning from long, unstructured texts by decomposing them into distinct knowledge aspects. This facilitates fine-grained alignment at both global and patch levels, while explicitly modeling medical concept relationships through ontology-guided mechanisms. We validate our framework in the field of dermatology, where comprehensive experiments demonstrate the effectiveness of each component. Our approach achieves state-of-the-art zero-shot performance on disease classification and cross-modal retrieval tasks across eight datasets. Our code and the augmented dataset Derm1M-AgentAug comprising over 400k skin-image-text pairs will be released upon acceptance.

%% file: section/2_introduction.tex
 Despite the success of deep learning in automating medical image analysis, supervised methods \cite{liu2020deep, chan2023histopathology} remain constrained by their reliance on extensive labeled datasets, which are costly and labor-intensive to curate. Vision-language pretraining (VLP) \cite{clip} offers a promising solution by aligning visual and textual information via large-scale image-text pairs. Recent medical VLP models \cite{monet,conceptclip,biomedica} have leveraged web-sourced pairs, ranging from educational textbooks and PubMed articles to YouTube videos, to achieve impressive zero-shot performance on diverse clinical tasks, such as disease classification and cross-modal retrieval.

However, existing VLP approaches utilizing web-crawled image-text pairs fail to fully capitalize on this rich knowledge due to two limitations. 

First, regarding \textit{\textbf{data quality}}, current datasets~\cite{derm1m,monet,quilt1m} are often characterized by substantial noise and information sparsity~\cite{10.1007/978-3-032-05169-1_3}, frequently failing to capture the diagnostic and morphological details visible in the images. For instance, as shown in Fig.~\ref{fig:MAGEN_example}, textbooks and PubMed articles typically provide only brief captions with diagnosis labels or morphological observations, while detailed diagnostic analysis resides in the main text, which is routinely ignored by automated extraction pipelines. YouTube videos introduce an additional challenge as extraction methods sample isolated keyframes from educational videos but associate them with video-level descriptions that often correspond to different content or temporal contexts, resulting in caption-image misalignment. These data quality limitations underscore significant potential for improvement in web-crawled medical datasets.

\begin{figure}[!t]
\begin{center}
{\includegraphics[width=1\linewidth]
{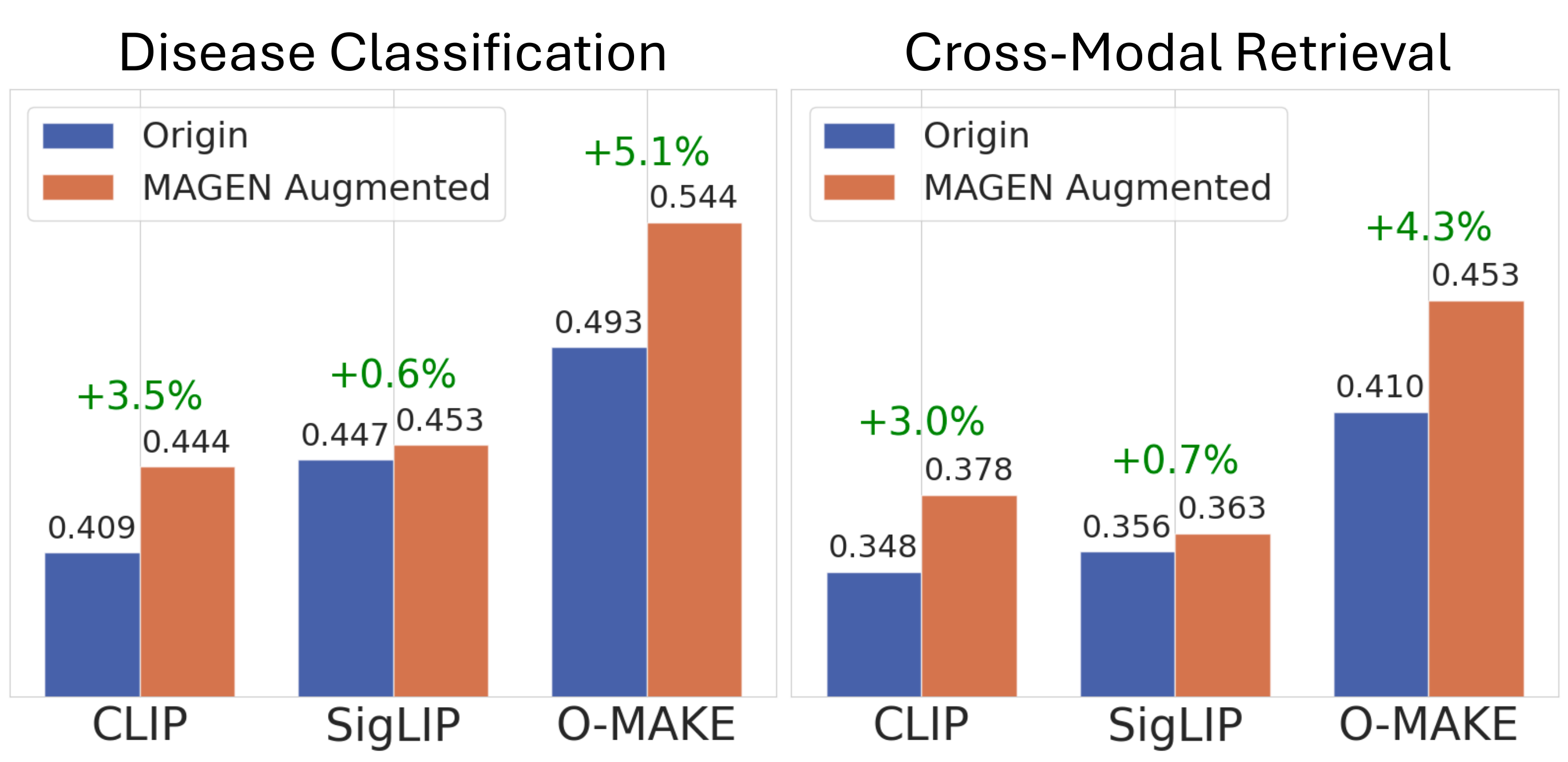}}
\end{center}
\vspace{-4mm}
\caption{\small MAGEN consistently improves zero-shot performance across diverse VLP approaches. Left: disease classification accuracy averaged across datasets. Right: cross-modal retrieval. All models trained on Derm1M show substantial gains when augmented with MAGEN-generated captions.}
\label{fig:caption_performance_comparasion}
\vspace{-3mm}
\end{figure}

Second, regarding \textit{\textbf{modeling}}, even when image-text pairs contain comprehensive medical knowledge, current VLP approaches~\cite{siglip,clip,coca} struggle to effectively leverage such information. Web-crawled image-text pairs in pretraining datasets often contain long, unstructured clinical descriptions that exceed the text encoder input limits of vision-language models (VLMs), resulting in truncation of diagnostically critical information. Beyond this length constraint, the fundamental challenge lies in effectively modeling the complex, multifaceted knowledge embedded within clinical descriptions through appropriate optimization objectives and VLP framework designs.

Existing solutions address these challenges separately. For data quality enhancement, general-domain recaptioning methods~\cite{whatif, hqclip, alip} leverage Multimodal Large Language Models (MLLMs) but suffer from domain gaps due to insufficient medical expertise, while medical-specific approaches like PathGen~\cite{pathgen} employ multi-agent systems with trained revision agents that limit generalization beyond training diseases. For knowledge modeling, recent methods have explored extended text encoders~\cite{longclip} and refined optimization objectives~\cite{laclip}, yet lack systematic frameworks that jointly address multi-aspect knowledge decomposition and disease relationship modeling. These limitations motivate our framework that addresses both challenges through integrated data enhancement and knowledge-guided pretraining.

In this paper, we propose a pretraining pipeline that effectively addresses limitations in both data quality and model optimization. First, to tackle the scarcity of high-quality textual supervision, we introduce the \textbf{Multi-Agent data GENeration (MAGEN)} system. MAGEN automatically synthesizes knowledge-rich captions for images that lack sufficient description. Specifically, it employs a collaborative workflow: a Captioning Agent utilizes a finetuned multimodal large language model and an off-the-shelf foundation model to generate precise clinical descriptions, while a Verification Agent leverages knowledge base retrieval to fact-check and correct morphological features and diagnostic findings. This multi-agent mechanism ensures the generation of accurate annotations while mitigating the risk of hallucinations. To validate MAGEN's effectiveness, we train mainstream VLP methods (CLIP, SigLIP) and our proposed framework (detailed below) on both original and MAGEN-enhanced datasets. As shown in Fig.~\ref{fig:caption_performance_comparasion}, MAGEN-enhanced data consistently improve zero-shot classification and retrieval performance across all methods.

Second, to fully exploit these knowledge-rich image–text pairs, we propose \textbf{Ontology-based Multi-Aspect Knowledge-Enhanced (O-MAKE)}. This pretraining strategy decomposes unstructured medical texts into distinct clinical aspects to facilitate granular alignment and relationship modeling. \textbf{1) Decomposition \& Alignment:} We decompose long descriptions via two complementary methods: splitting original captions and using LLMs to extract aspect-specific knowledge (e.g., morphological features). This enables multi-level alignment: at the \textit{global level}, images align with multiple explicit knowledge representations, circumventing context length constraints; at the \textit{patch level}, specific sub-captions are associated with diagnostically relevant regions, enabling the model to jointly characterize salient visual features. \textbf{2) Ontology-Guided Learning:} Inspired by clinical diagnostics, O-MAKE incorporates ontology-guided mechanisms at two levels. At the \textit{intra-sample level}, adaptive weighting prioritizes diagnostically critical aspects for each image. At the \textit{inter-sample level}, ontology-aware soft-label learning facilitates systematic knowledge sharing among hierarchically related diseases, enabling the model to exploit common morphological patterns across conditions. 

In this paper, our main contributions are summarized below:

1) We propose MAGEN, a multi-agent data generation system that automatically generates knowledge-enriched descriptions for images with insufficient annotations.

2) We propose O-MAKE, an ontology-based multi-aspect knowledge alignment framework that enables VLMs to learn from knowledge-rich image-text pairs through fine-grained alignment at both global and patch levels.

3) We introduce ontology-guided mechanisms that model relationships among medical knowledge within individual samples and across hierarchically related diseases, enabling VLMs to effectively exploit shared knowledge.

4) We validate our framework on the dermatology domain, a demanding domain with subtle visual distinctions and complex disease hierarchies. Our method achieves state-of-the-art performance across zero-shot diagnosis, long tail classification, and cross-modal retrieval across eight datasets.

Our preliminary MICCAI2025 work (MAKE)\cite{make} demonstrated that multi-aspect knowledge integration enhances medical vision-language pretraining for dermatological applications. This paper extends MICCAI2025\cite{make} into a comprehensive framework that addresses both data quality issues and suboptimal knowledge utilization through multi-agent data generation and modified ontology-enhanced multi-knowledge pretraining. Our main contributions include: (1) We propose a Multi-Agent Data Generation (MAGEN) system that transforms image-text pairs with poor textual descriptions into knowledge-enriched descriptions through foundation model-guided captioning and RAG-based verification; (2) We enhance the original multi-aspect knowledge-enhanced learning by employing soft-label learning, enabling VLMs to capture shared knowledge patterns across related samples with similar hierarchical disease relationships; (3) We systematically evaluate each MAGEN component and O-MAKE pretraining module through ablation studies, demonstrating progressive improvements as components are integrated; (4) We show that MAGEN-enhanced data consistently improves performance across representative VLP frameworks (CLIP, SigLIP, and O-MAKE) on both zero-shot disease classification and cross-modal retrieval tasks; (5) We provide in-depth analysis of our method's effectiveness through t-SNE visualizations and qualitative examples.

%% file: section/3_relatedwork.tex
\textbf{Vision-Language Pretraining.} CLIP~\cite{clip} pioneered vision–language pretraining (VLP) by leveraging billions of web-crawled image–text pairs, demonstrating strong zero-shot capabilities in classification, retrieval, and multimodal reasoning. Its success has catalyzed numerous medical VLP adaptations, which can be broadly categorized into two groups. General medical models, such as PMC-CLIP~\cite{pmcclip}, BiomedCLIP~\cite{biomedclip}, PubMedCLIP~\cite{pubmedclip}, and BMC-CLIP~\cite{biomedica}, are trained on image–text pairs derived from PubMed articles and biomedical literature to support biological and clinical interpretation across diverse imaging modalities. In contrast, specialty-specific models, such as PLIP~\cite{plip}, CPath-CLIP~\cite{sun2025cpath}, OphCLIP~\cite{ophclip}, and DermLIP~\cite{derm1m}, target specialized domains using focused, domain-curated datasets. Despite their differences, both approaches rely on large-scale web-collected data, which inherently introduces substantial noise and information sparsity.

\textbf{Augmented Image–Text Datasets.} To mitigate the noise in web-crawled data, recent research has focused on data augmentation strategies, particularly through synthetic recaptioning. General-domain methods like WhatIf \cite{whatif}, HQ-CLIP \cite{hqclip}, and ALIP \cite{alip} employ MLLMs to generate visually grounded, descriptive captions. However, these approaches face significant domain gaps when applied to medical imaging due to the insufficient clinical expertise of general-purpose models. While medical-specific efforts like PathGen~\cite{pathgen} address this by deploying MLLM-based multi-agent systems for pathology recaptioning, their verification relies on training a revision agent with GPT-4-generated instruction data. Consequently, their verification pipeline depends on disease patterns and visual features within the training distribution, lacking external knowledge integration to validate descriptions beyond the training domain.

\textbf{Knowledge-Enhanced Vision–Language Pretraining.} Integrating structured medical knowledge into VLP frameworks has emerged as a promising direction to improve semantic alignment. Existing methods primarily follow two paradigms: 1) Implicit Injection: Approaches like KEP \cite{kep} and KEEP \cite{keep} incorporate medical terminologies and relationships implicitly by enhancing the text encoder's representation space. 2) Concept-based Alignment: Methods such as ConceptCLIP \cite{conceptclip}, MedKLIP \cite{medklip}, and KAD \cite{kad} integrate knowledge during the alignment process by extracting and matching specific medical entities. Distinguishing our work from these approaches, our framework enables VLMs to model knowledge at two complementary levels. At the intra-sample level, we prioritise diagnostically relevant knowledge within individual descriptions through ontology-guided weighting. At the inter-sample level, we exploit the fact that hierarchically related diseases share common morphological features and diagnostic patterns, such as erythema and edema in inflammatory conditions, using these shared characteristics to provide enriched supervision signals across similar disease samples.

%% file: section/4_method.tex
\subsection{Overview}
Our framework consists of two key components. First, \textbf{Multi-Agent data GENeration (MAGEN)} automatically transforms quality-limited image-text pairs into knowledge-enriched descriptions via tool-based agent collaboration and RAG-based verification, producing an augmented dataset with enhanced textual quality. Second, \textbf{Ontology-Based Multi-Aspect Knowledge-Enhanced (O-MAKE)} pretraining framework leverages this augmented dataset through multi-aspect knowledge decomposition and ontology-guided learning. O-MAKE employs LLMs to extract and align distinct knowledge aspects with images, while modeling knowledge relationships through diagnostic adaptive weighting within samples and soft-label learning across hierarchically related diseases.

\begin{figure*}[!t]
\begin{center}
{\includegraphics[width=1\linewidth]
{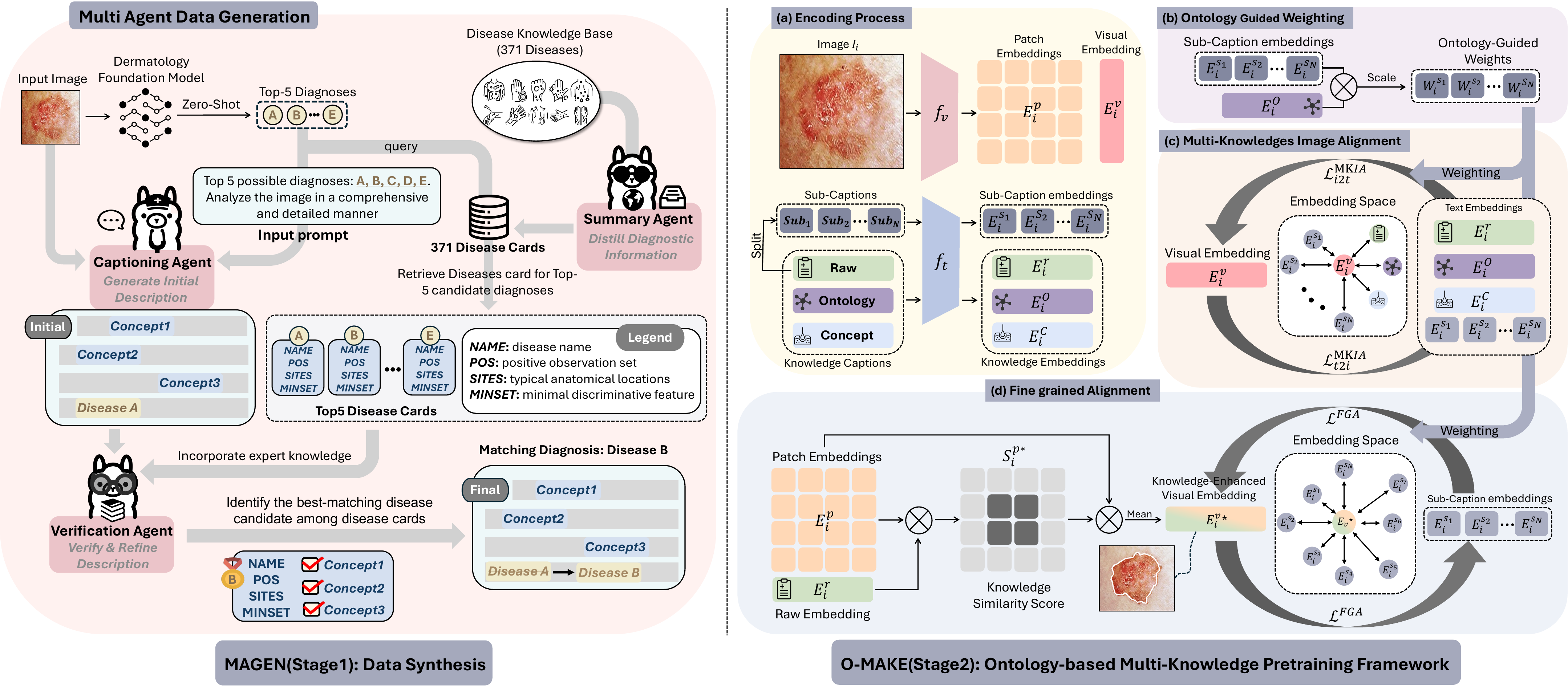}}
\end{center}
\vspace{-4mm}
\caption{\small
\textbf{Left:} Multi-Agent Data Generation (MAGEN) system with three key components: Captioning Agent (finetuned MLLM), Summary Agent, and Verification Agent. \textbf{Right:} Ontology-based multi-knowledge pretraining framework with (a) multi-aspect knowledge encoding, (b) ontology-guided weighting via similarity computation, (c) multi-knowledge image alignment ($\mathcal{L}^{MKIA}_{i2t}$ and $\mathcal{L}^{MKIA}_{t2i}$), and (d) fine-grained alignment ($\mathcal{L}^{FGA}$).}
\label{fig:Pretraining_Method}
\end{figure*}

\subsection{Multi-Agent Data Generation System}

\textbf{Pipeline:} As shown in Fig.\ref{fig:Pretraining_Method}-left, the system workflow comprises four stages. 1) The \textit{Dermatology Foundation Model} analyzes the input image to give top-5 differential diagnoses, providing a diagnosis prior. 2) The \textit{Captioning Agent} then generates a preliminary clinical description by leveraging both the image and these diagnostic priors. 3) Concurrently, the \textit{Summary Agent} queries a curated medical knowledge base for the top-5 candidates, generating structured summaries termed "Disease Card" that highlight key diagnosis criteria. 4) The \textit{Verification Agent} performs the final quality control. It triangulates information from the input image, the preliminary description, and the Disease Cards. Specifically, it checks the morphological claims in the description against the visual evidence and the clinical standards defined in the Disease Cards. It then rectifies any inconsistencies, producing a final description characterized by high-quality description and verified diagnostic correctness.

\textbf{Dermatology Foundation Model:}
To predict differential diagnosis from dermatological images, we employ a dermatology vision-language foundation model PanDermv2\cite{pandermv2}. The model uses 371 skin diseases from the Derm1M\cite{derm1m} ontology tree as the predefined disease pool. For a given image, the model encodes it into a visual embedding and performs zero-shot classification by computing cosine similarity with disease class text embeddings. This produces the top-5 candidate diagnoses, which serve as diagnostic priors for subsequent agents.

\textbf{Captioning Agent.} Serving as the core, the Captioning Agent is instantiated as a Multimodal Large Language Model (MLLM) based on the LLaVA framework. It is tasked with synthesizing an initial clinical description containing detailed morphological observations and a preliminary diagnostic assessment, conditioned on the dermatological image and diagnostic priors. To bridge the domain gap often found in general-purpose MLLMs, we initialize the agent with the dermatology foundation model PanDerm V2~\cite{panderm, pandermv2} as the vision encoder and Qwen3-14B~\cite{qwen3} as the LLM decoder, ensuring domain-specific visual representation and text generation. Following the LLaVA-v1.5~\cite{llava1.5} paradigm, we train the agent in two stages: (1) Visual-Language Alignment on the Derm1M~\cite{derm1m} dataset to align visual features with medical terminology; and (2) Instruction Fine-tuning on Derm1M-Instruct, a curated set of 98,460 captioning instructions derived from the original Derm1M dataset~\cite{derm1m}. This instruction set removes image-irrelevant text information and enriches descriptions with morphological details and diagnostic reasoning by querying GPT-4o.  During both training and inference, the agent is conditioned on the top-5 differential diagnoses predicted by the upstream Dermatology Foundation Model. This injection of diagnostic priors guides the agent to focus on disease-relevant features, thereby enhancing diagnostic accuracy.

\textbf{Summary Agent.} To enable knowledge-grounded verification, we construct a structured disease knowledge base using the Summary Agent. For each of the 371 diseases in the Derm1M ontology tree, we systematically collect comprehensive medical information via Google AI search, manually querying disease names and extracting returned content, averaging approximately 600 tokens per disease. We then employ Qwen2.5-72B-Instruct~\cite{qwen2.5} as the Summary Agent to distill these lengthy profiles into concise, structured Disease Cards of 60-120 tokens. Each card contains four diagnostically critical fields: \textit{NAME} (standardized disease name), \textit{POS} (positive observation set describing typical clinical features), \textit{SITES} (typical anatomical locations), and \textit{MINSET} (minimal discriminative features for differential Disease Cards). Consider the Disease Card for guttate psoriasis: \textit{NAME}: guttate psoriasis; \textit{POS}: small, red, scaly, drop-like spots; rapid onset; triggered by strep infection; \textit{SITES}: upper trunk, arms, legs, scalp; \textit{MINSET}: small, red, scaly spots; rapid onset. This structured compression enables efficient retrieval while retaining essential diagnostic criteria.

\textbf{Verification Agent.} To rectify potential hallucinations and refine clinical descriptions, we deploy Qwen2.5-VL-72B-Instruct~\cite{qwen2.5-vl} as the Verification Agent within a Retrieval-Augmented Generation (RAG) framework. This agent operates by triangulating information from three sources: the input image, the preliminary caption from the Captioning Agent, and the retrieved Disease Cards corresponding to the top-5 diagnostic candidates. We instruct the agent to perform a four-step reasoning process: (1) extract morphological claims from the preliminary caption; (2) cross-reference these claims against both the visual evidence and the criteria in the Disease Cards (POS, SITES, MINSET); (3) determine the best-matching diagnosis based on visual-textual consistency; and (4) synthesize a refined clinical description with verified diagnostic labels. To prevent forced misdiagnoses on out-of-distribution or ambiguous samples, the agent is configured to output "No definitive diagnosis" if no candidate demonstrates sufficient visual alignment.
 
\textbf{Dataset Augmentation with MAGEN.} To achieve a trade-off between efficiency and data quality, we apply MAGEN primarily to enhance low-quality image-text pairs in the pretraining dataset. We use DermLIP-PanDerm~\cite{derm1m} to compute cosine similarity between image and text embeddings, identifying pairs with low semantic alignment (similarity scores below 0.7) for recaptioning. MAGEN processes these pairs to generate knowledge-enriched descriptions with verified disease labels. For pairs flagged as "No definitive diagnosis" by the Verification Agent, we retain the initial descriptions from the Captioning Agent to avoid introducing erroneous information. This selective enhancement strategy produces an augmented dataset combining MAGEN-generated captions with original high-quality pairs.

\subsection{Multi-Aspect Knowledge-Enhanced VLP Framework}

As shown in Fig.~\ref{fig:Pretraining_Method}-right, our O-MAKE VLP framework comprises four core components: \textit{Ontology Guided Weighting}, \textit{Multi-Knowledge Image Alignment}, \textit{Fine-Grained Alignment}, and \textit{Ontology-Based Multi-knowledge Contrastive Learning}. We detail each of them below.

\textbf{Encoding Stage.} Our dataset comprises $M$ image-text pairs $(I_i, T_i^{r})$ for $i \in \{1, ..., M\}$, where $I_i$ is the image and $T_i^{r}$ is the raw caption. For each sample, we extract multiple textual representations to capture distinct knowledge aspects. Using an LLM, we extract two knowledge-specific captions from raw caption $T_i^{r}$: (1) \textit{Ontology Caption} $T_i^{o}$ containing the hierarchical disease path of disease label $d_i$ (extracted from $T_i^{r}$ and mapped to the ontology tree defined by dataset), and (2) \textit{Visual Concept Caption} $T_i^{c}$ emphasizing morphological attributes. These form a knowledge caption set $\{T_i^{r}, T_i^{o}, T_i^{c}\}$. To fully leverage semantic information in the raw caption, we further split the raw caption $T_i^{r}$ into $N$ sub-captions by sentences, yielding $\{T_i^{\text{sub}_j}\}^N_{j=1}$ where each \textit{Sub-Caption} corresponds to one sentence.

As shown in Fig.~\ref{fig:Pretraining_Method}-right(a), for the $i$-th sample, we encode the image through a vision encoder $f_v$ to obtain \textit{patch embeddings} $E_i^p=f_v(I_i)$, which are then aggregated into a global \textit{visual embedding} $E_i^v$ via mean pooling across patches.
The text encoder $f_t$ processes the knowledge caption set $\{T_i^{r}, T_i^{o}, T_i^{c}\}$ and sub-captions $\{T_i^{\text{sub}_j}\}_{j=1}^N$ to produce $3+N$ \textit{textual embeddings}:
\begin{equation}
E_i^t =\{(E_i^r, E_i^o, E_i^c), \{E_i^{\text{s}_j}\}_{j=1}^N\}
\label{equ:text_embeddings}
\end{equation}
comprising 3 \textit{knowledge embeddings} and $N$ \textit{sub-caption embeddings}. This approach not only accepts long captions exceeding the length limitation as text input, but also explicitly captures multi-aspect clinical knowledge within the caption.

\textbf{Ontology-Guided Weighting.} By mimicking how dermatologists prioritize clinical information based on diagnostic relevance, we introduce an ontology-guided weighting mechanism that adaptively assigns importance to each sub-caption. The ontology embedding $E_i^o$, which encodes disease-specific knowledge, serves as a semantic anchor to guide this prioritization process.

As illustrated in Fig.~\ref{fig:Pretraining_Method}-right(b), for the $i$-th sample, we compute the \textit{ontology-guided weights} for the sub-captions by measuring semantic similarity between sub-caption embeddings $\{E_i^{s_j}\}_{j=1}^{N}$ and the ontology embedding $E_i^o$:
\begin{equation}
\label{equ:ontology_guided_weight}
w_i = \frac{\{E_i^o \cdot E_i^{s_j}\}_{j=1}^{N}}{\max(\{E^o_i \cdot (E^{s_j}_i)\}_{j=1}^{N})}
\end{equation}
This yields a normalized weight vector $w_i = [w_i^1, \ldots, w_i^N]$ for the $i$-th sample's $N$ sub-captions, where diagnostically relevant sub-captions receive higher weights. These weights then guide subsequent alignment loss to prioritize diagnostically relevant correspondences, ensuring that clinically critical information receives greater emphasis.

\begin{figure}[!t]
\begin{center}
{\includegraphics[width=1\linewidth]
{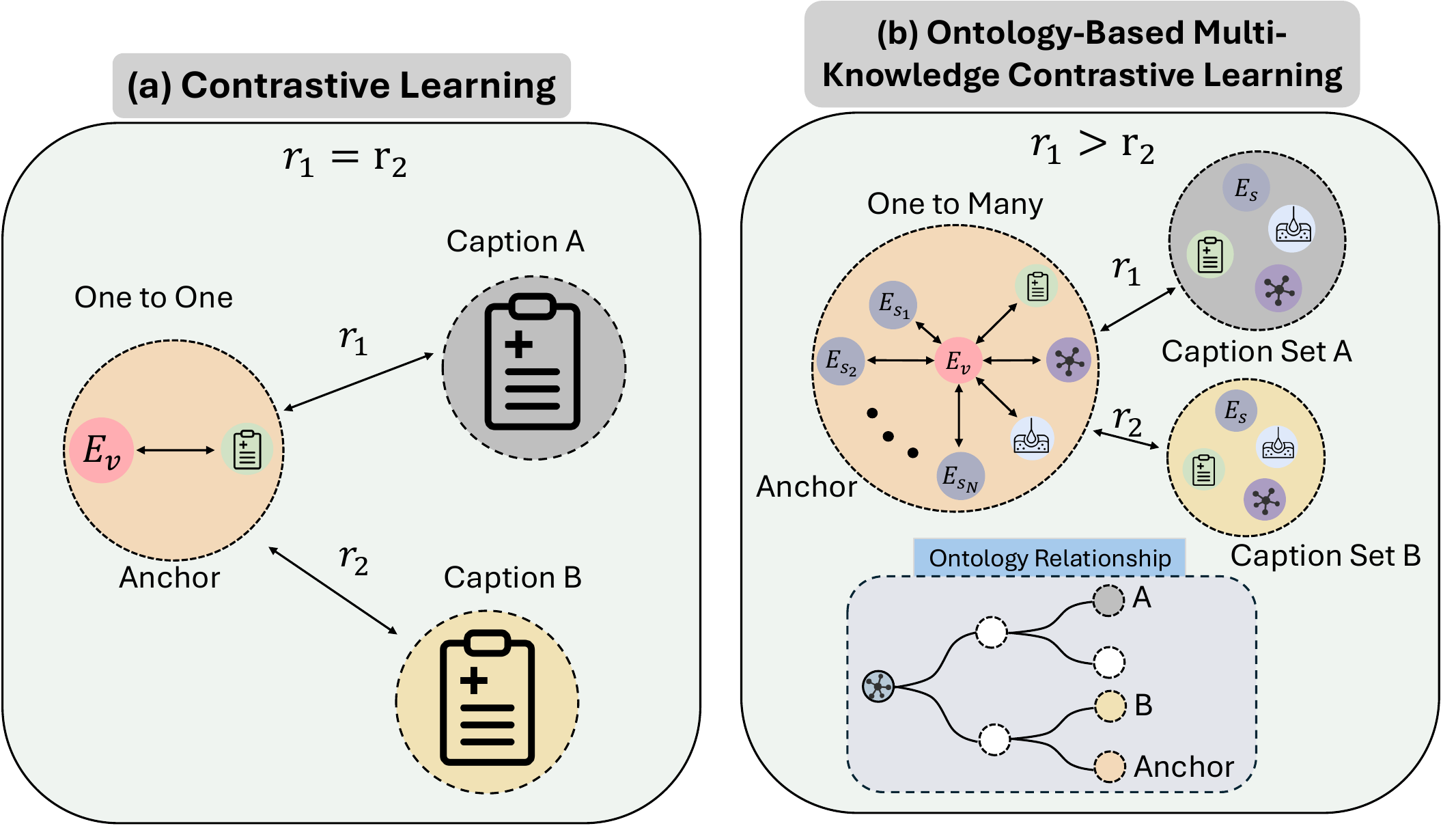}}
\end{center}
\vspace{-4mm}
\caption{\small Comparison between (a) traditional contrastive learning and (b) our ontology-based multi-knowledge approach.}
\label{fig:ontology}
\end{figure}

\textbf{Multi-Knowledge Image Alignment.} We adopt \textit{multi-positive contrastive learning}~\cite{DreamLIP,laclip,stablerep} to align multiple knowledge-specific representations with images as shown in Fig.~\ref{fig:Pretraining_Method}-right(c). While traditional contrastive learning aligns each image with a single text, multi-positive learning creates one-to-many alignment where each image forms $(N+3)$ positive alignment with all textual embeddings from Eq.~\ref{equ:text_embeddings}.

Specifically, we enforce \textit{Multi-Knowledge Image Alignment Loss} $\mathcal{L}^{MKIA}$ at two complementary levels: \textit{(1) Knowledge level} aligns the visual embedding with three knowledge embeddings $\{E_i^{r},E_i^{o},E_i^{c}\}$, and \textit{(2) Sub-caption level} aligns the visual embedding with $N$ sub-caption embeddings $\{E_i^{s_j}\}_{j=1}^{N}$, weighted such that subcaptions with higher diagnostic relevance contribute more to the loss. Given a batch of $B$ samples, the image-to-text alignment loss is:

\begin{equation}
\begin{aligned}
\mathcal{L}^{MKIA}_{i2t} =&
\underbrace{\frac{1}{B}\sum_{i=1}^{B}\sum_{k \in \{r,o,c\}} \mathcal{L}^{align}(E_i^v, E_i^k)}_{\text{Knowledge-level alignment}} \\
&+ \underbrace{\frac{1}{B}\sum_{i=1}^{B}\sum_{j=1}^{N} w_i^j \cdot \mathcal{L}^{align}(E_i^v, E_i^{s_j})}_{\text{Sub-caption level alignment}},
\end{aligned}
\label{equ:mkia_i2t}
\end{equation}

where for $i$-th sample: $E_i^{v}$ is the visual embedding, $E_i^{k}$ are knowledge embeddings ($k\in\{r,o,c\}$ for raw, ontology, and concept aspects), $E_i^{s_j}$ is the $j$-th sub-caption embedding, and $w_i^j$ are ontology-guided weights from Eq.~\ref{equ:ontology_guided_weight}. 

The symmetric text-to-image loss $\mathcal{L}^{MKIA}_{t2i}$ is computed analogously. We combine both directions as:
\begin{equation}
\mathcal{L}^{MKIA} = \frac{1}{2}(\mathcal{L}^{MKIA}_{i2t} + \mathcal{L}^{MKIA}_{t2i}).
\end{equation}
The alignment loss $\mathcal{L}_{\mathrm{align}}(\cdot,\cdot)$ will be detailed below.

\textbf{Ontology-Based Multi-Knowledge Contrastive Learning.} To define $\mathcal{L}^{align}$, we observe that medical conditions follow hierarchical taxonomies where ontologically related diseases share similar clinical features and diagnostic knowledge. However, traditional contrastive learning treats all negative samples uniformly. In a batch, for a given anchor image, only its corresponding text is considered a positive sample, while all other texts in the batch serve as negative samples. Traditional methods push the anchor equally far from all negatives, ignoring semantic relationships among them. As illustrated in Fig.~\ref{fig:ontology}-left, traditional contrastive learning maintains equal embedding distances to all negatives ($r_1 = r_2$), while our ontology-based approach encourages smaller distances to ontologically related diseases (Caption Set A) than to unrelated ones (Caption Set B), resulting in $r_1 > r_2$ as shown in Fig.~\ref{fig:ontology}-right.

Given a batch of $B$ image-text pairs with disease labels $\{d_i\}_{i=1}^{B}$, we construct an ontology-based similarity matrix $S \in \mathbb{R}^{B \times B}$, where each entry $S(i,j)$ quantifies the ontological relatedness between  $i$-th and $j$-th samples within a batch using the ontology tree provided by the dataset:
\begin{equation}
S(i,j) = \frac{2 \times |\text{shared\_path}(d_i, d_j)|}{|\text{path}(d_i)| + |\text{path}(d_j)|},
\label{equ:ontology_sim}
\end{equation}
where $\text{shared\_path}(d_i, d_j)$ denotes common ancestors between diseases $d_i$ and $d_j$ in the ontology tree, $\text{path}(d_i)$ is the path from the ontology tree root to disease $d_i$, and $S(i,j) \in [0,1]$ quantifies disease relatedness.

Traditional contrastive learning uses one-hot labels $y_{i,j}$ as supervision, where $y_{i,j} = 1$ for matching pairs ($i=j$) and $0$ otherwise, treating all negative samples uniformly. However, as ontologically related diseases share similar clinical knowledge, we construct soft labels by blending hard labels with ontological similarities:
\begin{equation}
\text{softlabel}(i,j) = (1-\beta) \cdot y_{i,j} + \beta \cdot \frac{\exp(S(i,j)/\tau_s)}{\sum_{k=1}^{B}\exp(S(i,k)/\tau_s)},
\label{equ:soft_label}
\end{equation}

Where $\beta$ is the hyper-parameter to control the influence of ontological structure. This formulation produces similarity-aware soft target distributions, assigning larger soft-label values to ontologically related diseases than to unrelated negatives.

To compute the alignment loss $\mathcal{L}_{align}$ between a visual embedding $E_i^v$ and a textual embedding $E_i^t$, where the textual embedding can be from any of the three knowledge aspects or the sub-captions, we calculate the predicted logits distribution across all batch samples:

\begin{equation}
p_{i,j}(E_i^v, E_j^t) = \frac{\exp(E_i^v \cdot E_j^t/\tau)}{\sum_{k=1}^{B}\exp(E_i^v \cdot E_k^t/\tau)}
\label{equ:pred_sim}
\end{equation}

Then, the alignment loss is defined as cross-entropy loss between the soft-label distribution and the predicted logits distribution:

\begin{equation}
\mathcal{L}^{align}(E_i^v, E_i^t) = -\sum_{j=1}^{B} \text{softlabel}(i,j) \log p_{i,j}(E_i^v, E_j^t)
\label{equ:ce}
\end{equation}

By minimizing this loss, the model not only aligns matching pairs but also learns from ontologically related samples, facilitating knowledge sharing across similar diseases.

\textbf{Fine-Grained Alignment: }To enhance fine-grained alignment, we draw inspiration from dermatologists who examine lesion regions from multiple knowledge aspects.
As shown in Fig.~\ref{fig:Pretraining_Method}-right(d), we align sub-caption embeddings with knowledge-enhanced patch representations to capture local visual-semantic correspondences.

Specifically, we first compute knowledge similarity scores between patch embeddings $E_i^p = [v_i^{p_1}, \dots, v_i^{p_{HW}}] \in \mathbb{R}^{HW \times d}$ and the raw text embedding $E_i^r \in \mathbb{R}^d$ via dot product:
\begin{equation}
z_i = [z_i^1, \dots, z_i^{HW}] = E_i^p \cdot E_i^r \in \mathbb{R}^{HW},
\end{equation}
where $z_i^n = v_i^{p_n} \cdot E_i^r$ represents the similarity score between the $n$-th patch embedding and the raw caption embedding.
These scores are then normalized and used to weight the patch embeddings, highlighting regions with strong semantic relevance.
The knowledge-enhanced visual embedding is computed as:
\begin{equation} 
\label{equ:knowledge_enhanced_embedding} 
E^k_i = \sum_{n=1}^{HW} v_i^{p_n} \cdot \frac{z_i^n}{\sum_{j=1}^{HW}z_i^j},
\end{equation}
where the weighted sum aggregates patch embeddings based on their knowledge relevance.

The fine-grained alignment loss aligns each sub-caption embedding with the knowledge-enhanced visual embedding:
\begin{equation}
\label{equ:slra}
\mathcal{L}_{FGA} = -\sum_{i=1}^{B}\sum_{j=1}^{N}\log\frac{\exp(\text{sim}(E^{s_j}_i, E_i^k) / \tau)}{\sum_{n=1}^{B}\exp(\text{sim}(E^{s_j}_i, E_n^k)/\tau)}
\end{equation}
where $E^{s_j}_i$ denotes the $j$-th sub-caption embedding of sample $i$, and $\text{sim}(\cdot,\cdot)$ computes cosine similarity.

Our final pretraining objective combines both global alignment and patch-level alignment:
\begin{equation}
\label{equ:total_loss}
\mathcal{L}_{total} = \mathcal{L}^{MKIA} + \lambda\mathcal{L}^{FGA},
\end{equation}
where $\lambda$ is a hyperparameter balancing the two objectives.

%% file: section/5_experiments.tex
\begin{table*}[t]
\centering
\scriptsize
\caption{Zero-shot performance comparison across datasets. 
Left: classification accuracy (ACC). Right: long-tail classification accuracy (ACC).}
\renewcommand{\arraystretch}{1.0}
\resizebox{\textwidth}{!}{%
\begin{tabular}{c|c|ccccc|c|cc|c}
\toprule
\multirow{3}{*}{\textbf{Category}} & \textbf{Model Name} &
\multicolumn{6}{c|}{\textbf{Disease Classification (ACC)}} &
\multicolumn{3}{c}{\textbf{Long Tail Classification (ACC)}} \\
\cline{2-2} \cline{3-8} \cline{9-11}
 & \textbf{Dataset} &
\textbf{PAD} & \textbf{F17K} & \textbf{SD-128} & \textbf{SNU-134} & \textbf{Daffodil} & \multirow{2}{*}{\textbf{Avg.}} &
\textbf{SD-Tails} & \textbf{SNU-Tails} & \multirow{2}{*}{\textbf{Avg.}} \\
& \textbf{Class Num} & \textbf{6} & \textbf{113} & \textbf{128} & \textbf{134} & \textbf{5} &  & \textbf{70} & \textbf{85} &  \\
\hline
\multirow{6}{*}{\makecell{\textbf{Open VLMs}$^{\text{\tiny$\clubsuit$}}$}}
 & CLIP-OPENAI     & 0.433 & 0.063 & 0.073 & 0.073 & 0.454 & 0.219 & 0.148 & 0.118 & 0.133 \\
 & MONET           & 0.474 & 0.150 & 0.217 & 0.150 & 0.758 & 0.350 & 0.311 & 0.179 & 0.245 \\
 & BMC-CLIP        & 0.526 & 0.107 & 0.137 & 0.140 & 0.682 & 0.318 & 0.205 & 0.175 & 0.190 \\
 & BioMedCLIP      & 0.430 & 0.089 & 0.132 & 0.097 & 0.589 & 0.267 & 0.192 & 0.136 & 0.164 \\
 & DermLIP-PanDerm & 0.615 & 0.319 & \underline{0.403} & 0.322 & \underline{0.799} & \underline{0.492} & 0.513 & \underline{0.419} & \underline{0.466} \\
 & DermLIP-ViTB16  & \underline{0.627} & 0.229 & 0.287 & 0.253 & 0.733 & 0.426 & 0.424 & 0.312 & 0.368 \\
\hline
\multirow{3}{*}{\makecell{\textbf{Retrained}$^{\text{\tiny$\spadesuit$}}$}}
 & SigLIP          & 0.582 & 0.283 & 0.362 & 0.313 & 0.728 & 0.453 & 0.443 & 0.397 & 0.420 \\
 & CoCa            & 0.556 & 0.236 & 0.315 & 0.240 & 0.646 & 0.399 & 0.401 & 0.279 & 0.340 \\
 & CLIP            & 0.571 & 0.283 & 0.340 & 0.286 & 0.740 & 0.444 & 0.408 & 0.357 & 0.383 \\
\hline

\rowcolor{gray!12}
\multicolumn{1}{c|}{\cellcolor{gray!12}} &
MAKE &
0.596 & \underline{0.324} & 0.395 & \underline{0.326} & 0.787 & 0.485 &
\underline{0.524} & 0.393 & 0.459 \\
\rowcolor{gray!12}
\multicolumn{1}{c|}{\multirow{-2}{*}{\cellcolor{gray!12}\textbf{Ours}$^{\text{\tiny$\spadesuit$}}$}} & 
\textbf{O-MAKE} &
\textbf{0.667} & \textbf{0.371} & \textbf{0.460} & \textbf{0.390} & \textbf{0.832} & \textbf{0.544} &
\textbf{0.558} & \textbf{0.457} & \textbf{0.508} \\

\bottomrule
\end{tabular}
}

\begin{minipage}{\textwidth}
\vspace{1mm}
\scriptsize
$^{\text{\tiny$\clubsuit$}}$ Open VLP models evaluated using publicly released checkpoints.\\
$^{\text{\tiny$\spadesuit$}}$ Mainstream VLP methods and our O-MAKE trained on the same MAGEN-enhanced pretraining dataset. \\
MAKE is our prior work published in MICCAI 2025, and we report results using its publicly released checkpoint.
\end{minipage}
\label{tab:zero_shot_classification}
\end{table*}

\subsection{Experimental Setup}

\textbf{Pretraining Dataset.} We utilize Derm1M~\cite{derm1m} as the foundational pretraining dataset. It comprises 403,563 dermatological image–text pairs sourced from diverse repositories, including PubMed articles, textbooks, and social media cases. The dataset is organized under a four-level hierarchical ontology comprising 130 clinical concepts and over 390 skin conditions. 

\textbf{Dataset Augmentation Details.} Starting from the Derm1M dataset containing 403,563 dermatological image-text pairs, we identify low quality pairs using DermLIP-PanDerm~\cite{derm1m} to compute image-text cosine similarity. Pairs with similarity scores below 0.7 are selected for MAGEN processing, resulting in 183,934 candidates (45\% of the dataset). After processing through MAGEN, 133,930 pairs receive knowledge-enriched descriptions with verified disease labels. The remaining 50,004 pairs are flagged as "No definitive diagnosis" by the Verification Agent and retain the Captioning Agent's initial descriptions to maintain data quality. 

\textbf{Evaluation Datasets:}
For evaluation, we use eight downstream datasets across three task categories: 
(1) \textit{Zero-shot disease classification}: PAD~\cite{PACHECO2020106221} for skin cancer, Fitzpatrick17K~\cite{groh2021evaluating}, SNU-134~\cite{Han2019}, and SD-128~\cite{10.1007/978-3-319-46466-4_13} for general skin conditions, and Daffodil~\cite{daffodil} for rare diseases; 
(2) \textit{Long-tail classification}: SNU-Tails, containing 85 classes with fewer than 15 samples each from SNU-134~\cite{Han2019}, and SD-Tails, containing 70 classes with fewer than 20 samples each from SD-198\cite{10.1007/978-3-319-46466-4_13}, both covering diverse general dermatological conditions with highly imbalanced class distributions;
(3) \textit{Cross-modal retrieval}: SkinCAP~\cite{skincap} for both image-to-text and text-to-image retrieval. 
Following CLIP~\cite{clip}, we perform zero-shot evaluation without further fine-tuning.

\textbf{Pretraining Details:}
Following CLIP~\cite{clip}, we adopt ViT-B/16~\cite{vit} as the image encoder and GPT-2~\cite{gpt2} with a context length of 77 as the text encoder. Our model is pretrained for 15 epochs with a batch size of 2048, a learning rate of 1e-4, and a 1500-step warmup with weight decay of 0.1. Images are processed at $224\times224$ resolution. We use the final checkpoint for all compared VLP approaches and perform extensive hyperparameter tuning to determine optimal configurations. The loss weighting factor $\lambda$ is set to 0.7. Following CLIP, both temperature parameters $\tau$ and $\tau_s$ are set to 0.07. The soft-label blending parameter $\beta$ is set to 0.05.

\subsection{Comparisons with Existing Methods}

We compare O-MAKE with two categories of vision-language models.

\textbf{Open VLMs.} This category includes general-domain models such as CLIP-OPENAI~\cite{clip}, biomedical VLMs like BioMedCLIP~\cite{biomedclip} and BMC-CLIP~\cite{biomedica} trained on large-scale biomedical data, and dermatology-specific models including MONET~\cite{monet}, which is trained on 105,550 dermatological PubMed and textbook captions, and DermLIP-PanDerm~\cite{derm1m}, which employs a dermatology-pretrained vision encoder (PanDerm v1\cite{panderm}) paired with a more capable text encoder capable of processing longer text sequences and is trained on the original Derm1M dataset. All models are evaluated using their publicly released checkpoints.

\textbf{Retrained VLP.} For fair comparison, we retrain mainstream VLP approaches including CLIP~\cite{clip}, SigLIP~\cite{siglip}, and CoCa~\cite{coca} on our MAGEN-augmented pretraining dataset.

\textbf{Zero-Shot Benchmark.} Table~\ref{tab:zero_shot_classification} presents the zero-shot performance comparison across disease classification, long tail classification. Our O-MAKE achieves superior performance across all task categories, demonstrating strong generalization capabilities.

\textit{1) Disease Classification.} O-MAKE achieves an average accuracy of 54.4\%, substantially outperforming all baselines, including our prior MICCAI work MAKE (48.5\%). Three key findings emerge: First, \textit{superior multi-class recognition:} On datasets with over one hundred disease categories, O-MAKE achieves 46.0\% on SD-128, 39.0\% on SNU-134, and 37.1\% on F17K. Compared to the best open baseline DermLIP-PanDerm, these represent improvements of 5.7\%, 6.8\%, and 5.2\% respectively, and compared to MAKE, improvements of 6.5\%, 6.4\%, and 4.7\%. Second, \textit{exceptional rare disease recognition:} O-MAKE achieves 83.2\% on the Daffodil rare disease benchmark, surpassing DermLIP-PanDerm by 3.3\% and MAKE by 4.5\%. Third, \textit{effectiveness of knowledge-enhanced learning}: O-MAKE achieves 9.1\% improvement over the best retrained baseline on the same dataset, validating our framework’s superior capacity for knowledge utilization.

\textit{2) Long-Tail Classification.} O-MAKE demonstrates robust performance on highly imbalanced distributions, achieving 50.8\% average accuracy.  On SD-Tails, O-MAKE reaches 55.8\%, surpassing DermLIP-PanDerm by 4.5\%, MAKE by 3.4\%, and the best retrained baseline SigLIP by 11.5\%.

\begin{table}[t]
\centering
\small
\caption{Retrieval performance comparison on SkinCap dataset. 
Average Recall@10/50/100 for Image-to-Text and Text-to-Image tasks.}
\vspace{-1mm}
\renewcommand{\arraystretch}{1.02}
\resizebox{\columnwidth}{!}{
\begin{tabular}{c|c|c|c|c}
\toprule
\textbf{Category} & \textbf{Model Name} & 
\textbf{Image-to-Text} & \textbf{Text-to-Image} & \textbf{Average} \\
\hline
\multirow{6}{*}{\textbf{General}} 
 & CLIP-OPENAI & 0.222 & 0.173 & 0.198 \\
 & MONET & 0.312 & 0.325 & 0.319 \\
 & BMC-CLIP & 0.373 & 0.375 & 0.374 \\
 & BioMedCLIP & 0.316 & 0.303 & 0.310 \\
 & DermLIP-PanDerm & 0.396 & \underline{0.417} & 0.407 \\
 & DermLIP-ViTB16 & 0.337 & 0.333 & 0.335 \\
\hline
\multirow{3}{*}{\makecell{\textbf{Retrained}}}
 & SigLIP & 0.353 & 0.373 & 0.363 \\
 & CoCa & 0.312 & 0.334 & 0.323 \\
 & CLIP & \underline{0.408} & 0.411 & \underline{0.410} \\
\hline

\rowcolor{gray!12}
\multicolumn{1}{c|}{\cellcolor{gray!12}} &
MAKE & 0.404 & 0.400 & 0.402 \\
\rowcolor{gray!12}
\multicolumn{1}{c|}{\multirow{-2}{*}{\cellcolor{gray!12}\textbf{Ours}$^{\text{\tiny$\spadesuit$}}$}} & 
\textbf{O-MAKE} & \textbf{0.450} & \textbf{0.456} & \textbf{0.453}
\\

\bottomrule
\end{tabular}}
\label{tab:retrieval}
\end{table}

\textit{3) Cross-Modal Retrieval.} Table~\ref{tab:retrieval} presents zero-shot retrieval performance on the SkinCAP dataset. O-MAKE achieves an average recall of 45.3\% across R@10/50/100 metrics, substantially outperforming both general models and retrained baselines. For image-to-text retrieval, O-MAKE achieves 45.0\%, surpassing the best open baseline DermLIP-PanDerm by 5.4\% and the best retrained baseline CLIP by 4.2\%. For text-to-image retrieval, O-MAKE achieves 45.6\%, outperforming DermLIP-PanDerm and CLIP by 3.9\% and 4.5\%, respectively. Notably, the SkinCAP dataset contains lengthy captions with an average of 80 tokens, exceeding the 77-token limit of both CLIP and O-MAKE text encoders. Despite this constraint, O-MAKE's superior results demonstrate that our multi-aspect knowledge decomposition and ontology-guided learning effectively capture and utilize critical information from truncated long clinical descriptions, enabling robust cross-modal alignment.

\subsection{Ablation Studies}
\begin{table}[t]
    \centering
    \small
    \caption{Ablation study on multi-agent data generation components. 
    Classification accuracy (ACC) is reported per dataset.}
    \vspace{-1mm}
    \renewcommand{\arraystretch}{1.12}
    \resizebox{\columnwidth}{!}{
    \begin{tabular}{c|c|c|ccccc|c}
        \toprule
        \multicolumn{3}{c|}{\textbf{Multi-Agent Pipeline}} &
        \multicolumn{6}{c}{\textbf{Datasets}} \vspace{0.5mm}\\
        \cline{1-9}
        \textbf{Caption} & \textbf{Tool} & \textbf{Verification} &
        \textbf{PAD} & \textbf{F17K} & \textbf{SD-128} & \textbf{SNU-134} & \textbf{Daffodil} & \textbf{Avg.} \\
        \hline
         &  &  & 0.558 & 0.332 & 0.399 & 0.349 & 0.825 & 0.493 \\
        \checkmark &  &  & 0.583 & 0.332 & 0.376 & 0.354 & 0.811 & 0.491 \\
        \checkmark & \checkmark &  & 0.652 & \textbf{0.371} & 0.416 & 0.369 & \textbf{0.846} & 0.531 \\
        \rowcolor{gray!20}
        \checkmark & \checkmark & \checkmark & \textbf{0.667} &\textbf{0.371} & \textbf{0.460} & \textbf{0.390} & 0.832 & \textbf{0.544} \\
        \bottomrule
    \end{tabular}}
    \label{tab:ablation_agent_acc}
\end{table}

\begin{table}[t]
\small
\centering
\caption{Ablation on pretraining components. ACC is reported per dataset. $^*$ denotes MKCL without using sub-captions.}
\vspace{-1mm}
\renewcommand{\arraystretch}{1.12}
\resizebox{\columnwidth}{!}{%
\begin{tabular}{l|ccccc|c}
\toprule
\textbf{Modules} &
\textbf{PAD} & \textbf{F17K} & \textbf{SD-128} & \textbf{SNU134} & \textbf{Daff.} & \textbf{Avg.} \\
\midrule
Baseline & 0.571 & 0.283 & 0.340 & 0.286 & 0.740 & 0.444 \\
M$^*$ & 0.588 & 0.344 & 0.413 & 0.307 & 0.705 & 0.472 \\
M & 0.600 & 0.350 & 0.414 & 0.317 & 0.773 & 0.491 \\
M + F & 0.646 & 0.355 & 0.455 & \textbf{0.395} & 0.786 & 0.527 \\
M + F + O & 0.658 & 0.363 & 0.448 & 0.381 & 0.828 & 0.535 \\
\rowcolor{gray!20}
M + F + O + W &
\textbf{0.667} & \textbf{0.371} & \textbf{0.460} & 0.390 & \textbf{0.832} & \textbf{0.544} \\
\bottomrule
\end{tabular}%
}
\label{tab:ablation_pretraining_singlecol_module}
\end{table}

\textbf{Contribution of Multi-Agent Components.} 
Table~\ref{tab:ablation_agent_acc} evaluates the individual contribution of each component in MAGEN through systematic ablation. We compare four configurations: (1) baseline using original captions of Derm1M, (2) recaptioning with the Captioning Agent (finetuned Captioning MLLM), (3) adding the dermatology foundation model PanDerm v2 to provide top-5 disease candidates as context, and (4) the complete pipeline with Verification Agent. For each configuration, we generate the corresponding dataset, retrain a VLM using O-MAKE, and evaluate zero-shot disease classification performance.

As shown in Table~\ref{tab:ablation_agent_acc}, the results reveal progressive improvements as components are added. The baseline trained using the original Derm1M dataset achieves 49.3\% average accuracy, while using the Captioning Agent alone does not improve the performance. A substantial improvement emerges when incorporating the foundation model, which boosts average performance to 53.1\%, representing a 3.8\% gain over baseline. This improvement is particularly pronounced on PAD (+9.4\%), SD-128 (+1.7\%), and Daffodil (+2.1\%). Adding the Verification Agent in the complete pipeline further increases average performance to 54.4\%, contributing an additional 1.3\%. These results show that the foundation model provides the most significant performance contribution, while the Verification Agent offers complementary benefits.

\textbf{Contribution of Pretraining Components.} 
Table~\ref{tab:ablation_pretraining_singlecol_module} systematically evaluates each O-MAKE component. We start from a CLIP baseline and progressively add: M (Multi-Knowledge Image Alignment using multi-positive contrastive loss following MAKE~\cite{make}), F (Fine-Grained Alignment), O (Ontology-based soft labels), and W (ontology-guided Weighting).

\begin{figure}[!t]
\centering{\includegraphics[width=\columnwidth]{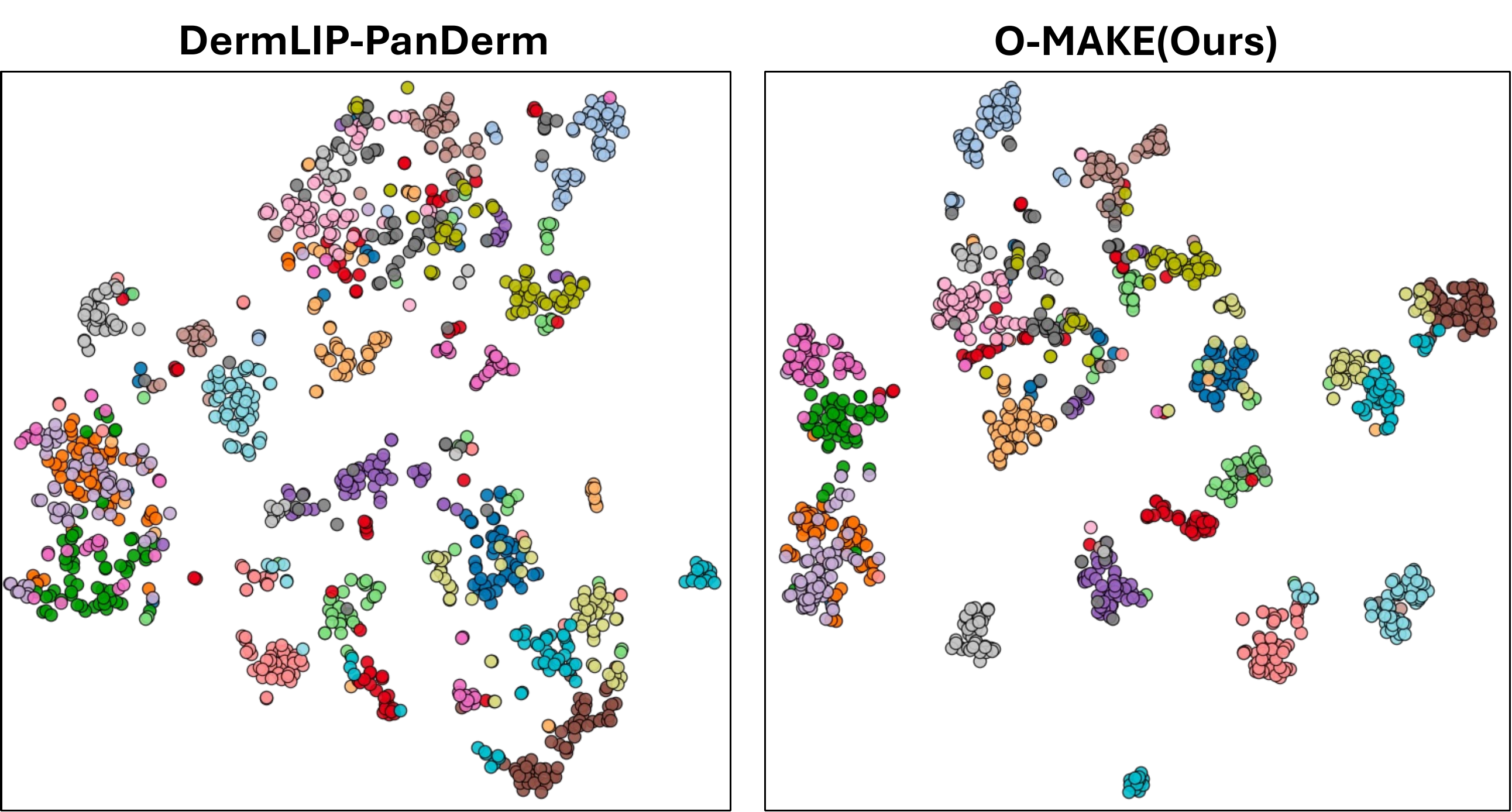}}
\caption{\textbf{T-SNE Visualization of Learned Visual Representations.} Comparison of image embeddings from vision encoders on the top-20 classes in SD128 dataset.}
\label{fig:tsne-visualization}
\end{figure}

Performance was improved as components were added, as shown in Table~\ref{tab:ablation_pretraining_singlecol_module}. From the CLIP baseline at 44.4\%, multi-knowledge alignment (M$^*$) improves performance to 47.2\%, a gain of +2.8\%. Adding subcaptions (M) further increases accuracy to 49.1\%, improving by an additional +1.9\%. Fine-grained alignment (M+F) provides the largest gain of +3.6\%, reaching 52.7\%, with particularly great improvements of +4.6\% on PAD and +4.1\% on SD-128. Introducing ontology-based soft labels (M+F+O) adds +0.8\% on average, achieving 53.5\%. Notably, the ontology component yields a pronounced +4.2\% improvement on Daffodil, validating that ontological relationships facilitate knowledge sharing for rare diseases. The complete framework with ontology-guided weighting reaches 54.4\%, yielding a final gain of +0.9\%. These results confirm that all components contribute positively, with fine-grained alignment and multi-knowledge learning being most impactful overall, while ontology-based learning particularly benefits rare disease scenarios.

\begin{figure*}[!t]
\centering{\includegraphics[width=1\linewidth]{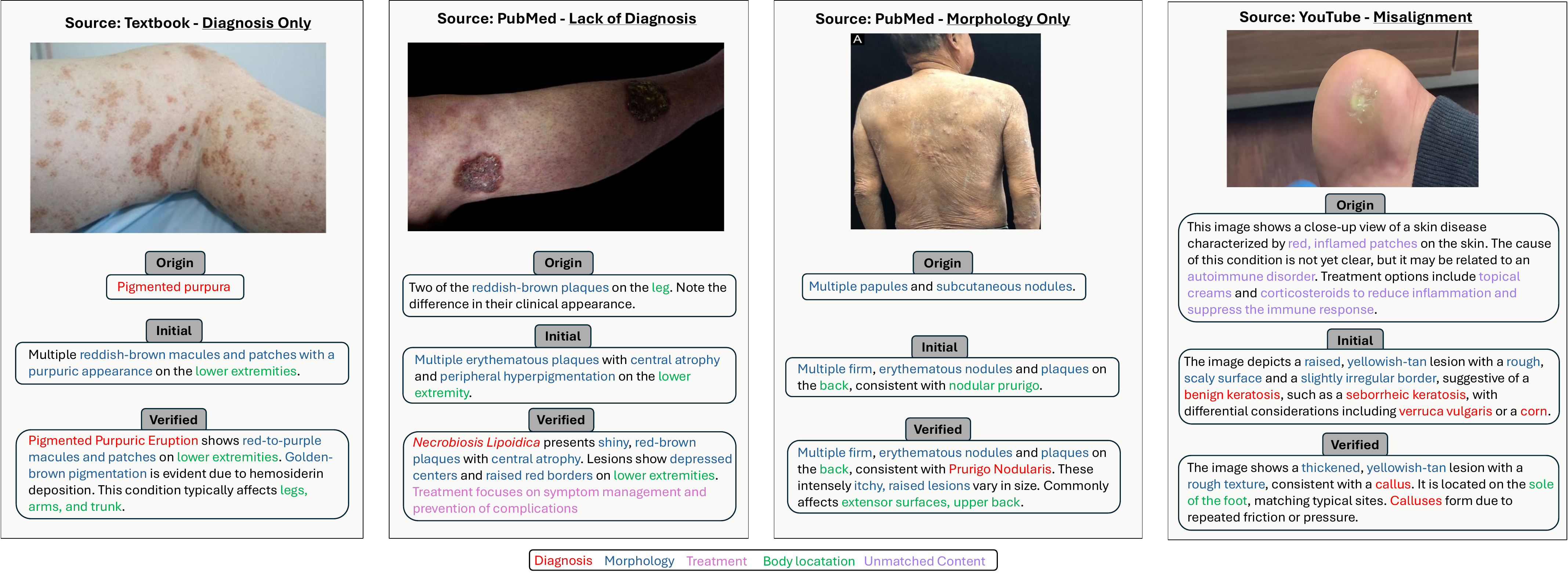}}
\caption{\textbf{Qualitative Examples of Multi-Agent Data Generation.} MAGEN transforms low-quality web-crawled captions (Origin) into knowledge-enriched descriptions through two stages: Captioning Agent generates morphology-focused descriptions guided by a foundation model(Initial), followed by Verification Agent's RAG-based refinement for accurate diagnoses (Verified).}
\label{fig:MAGEN_example}
\end{figure*}

\subsection{Analysis}

\textbf{Visualization of Vision Encoder Representations.} To further understand the quality of learned visual representations, we visualize image embeddings using t-SNE on the top-20 disease classes from the SD-128 dataset in Fig.\ref{fig:tsne-visualization}. We compare our O-MAKE framework against DermLIP-PanDerm\cite{derm1m}, which achieves the second-best performance on our zero-shot disease classification benchmark. As shown in the left panel of Fig.~\ref{fig:tsne-visualization}, DermLIP-PanDerm produces scattered clusters with substantial inter-class overlap, indicating difficulty in learning discriminative boundaries between similar dermatological conditions. In contrast, our O-MAKE framework (right panel) demonstrates significantly improved clustering with compact intra-class grouping and clear inter-class separation. This superior clustering structure validates that our pretraining framework enables the vision encoder to learn more discriminative feature representations.

\textbf{Qualitative Examples of Multi-Agent Data Generation.}
Fig.~\ref{fig:MAGEN_example} illustrates how MAGEN addresses quality issues in web-crawled dermatological image-text pairs. The original captions exhibit distinct limitations: a textbook case provides only a diagnosis label without morphological descriptions; the first PubMed case lacks a definitive diagnosis despite containing morphological observations; the second PubMed case contains only superficial morphology without diagnostic or anatomical information; and the YouTube case suffers from caption-image misalignment, describing "red, inflamed patches" that contradict the actual visual content.

Through MAGEN's multi-agent augmentation, these incomplete descriptions are systematically enriched. For the textbook case, MAGEN supplements the diagnosis label with comprehensive morphological observations and precise anatomical distributions. For the first PubMed case, the vague morphological description is transformed to include definitive disease identification and relevant treatment information. For the second PubMed case, MAGEN establishes an accurate diagnosis while enriching the text with detailed lesion characteristics and clinical symptoms. For the misaligned YouTube case, MAGEN corrects the inconsistent description, replacing erroneous inflammatory features with accurate morphological observations and a diagnosis matching the visual content.

In all cases, the verified captions integrate multiple knowledge aspects: accurate diagnoses (red), detailed morphological features (blue), anatomical distributions (green), and treatment information (pink). This demonstrates MAGEN's capability to identify and compensate for different types of knowledge gaps, converting fragmented, single-aspect descriptions into comprehensive clinical narratives that provide richer training signals for vision-language alignment.

%% file: section/6_discussion.tex
In this work, we addressed two critical challenges in medical vision-language pretraining: low-quality web-crawled data and ineffective knowledge utilization from long medical descriptions. We proposed a pretraining pipeline integrating Multi-Agent data GENeration (MAGEN) with Ontology-based Multi-Aspect Knowledge-Enhanced (O-MAKE) pretraining. MAGEN systematically transforms low-quality image-text pairs into knowledge-enriched descriptions through foundation model-assisted captioning and RAG-based verification, eliminating manual annotation requirements while ensuring high-quality textual content. As demonstrated in Fig.~\ref{fig:caption_performance_comparasion}, MAGEN-enhanced data consistently improves zero-shot classification and retrieval performance across diverse VLP methods, validating its generalizability for augmenting medical image-text datasets. O-MAKE addresses the challenge of modeling knowledge in long text through multi-aspect decomposition and ontology-guided learning, enabling systematic knowledge transfer across hierarchically related diseases. Comprehensive experiments demonstrate state-of-the-art zero-shot performance on disease classification, long-tail recognition, and cross-modal retrieval benchmarks.
While validated on dermatology, our modular framework can be readily extended to other medical domains. Future directions include adapting this approach to additional specialties and general medicine, incorporating multimodal clinical data beyond images and text, and exploring dynamic ontology construction for emerging diseases.